\UseRawInputEncoding
\documentclass[letterpaper, 10 pt, conference]{ieeeconf}  % Comment this line out if you need a4paper

\IEEEoverridecommandlockouts                              % This command is only needed if 
\usepackage{longtable}% for long tables
\usepackage{booktabs}
\usepackage{algorithm}
\usepackage{bbm}
\usepackage{xcolor}
\usepackage{subfigure}
\usepackage{amsfonts}
\usepackage{amsmath,amssymb,amsfonts}
\usepackage{algorithmic}
\usepackage{graphicx}
\usepackage{textcomp}
\usepackage{xcolor}
\usepackage{subfigure}
\usepackage{makecell}
\usepackage{cite}
\usepackage{graphicx}
\usepackage{amsmath}
\usepackage{amssymb}
\usepackage{booktabs}
\usepackage{multicol}
\usepackage{multirow}
\usepackage{bm}
\usepackage{comment}
\usepackage{booktabs}
\usepackage{array}
\usepackage{hyperref}
\hypersetup{hidelinks}
\usepackage{cleveref} 
\usepackage{booktabs}
\usepackage{array} % 提供了扩展的列定义选项，如 m for vertical centering
\usepackage{siunitx} % 使用siunitx来对齐数字
\Crefformat{figure}{#2Fig.~#1#3}
\Crefmultiformat{figure}{Figs.~#2#1#3}{ and~#2#1#3}{, #2#1#3}{ and~#2#1#3}
\begin{document}
\title{\LARGE \bf
%Motion-Aware Dynamic Latent Diffusion Model for Tele-operated Robot Video Debluring
UAV Video Deblurring via Motion-Aware Diffusion: A Path to Robust Target Detection
{  
}}
\author{Zhiqiang HU, Shouren HUANG and Masatoshi ISHIKAWA% <-this % stops a space
\thanks{}% <-this % stops a space
\thanks{The authors are with the Research Institute for Science \& Technology, Tokyo University of Science
        {\tt\small \{zhiqiang.hu, huang, ishikawa\}@ishikawa-vision.org}}%
}

\maketitle
\thispagestyle{empty}
\pagestyle{empty}

\begin{abstract}
Unmanned Aerial Vehicles (UAVs) play a crucial role in various scenarios ranging from disaster response to traffic surveillance. However, aerial video footage often suffers from severe motion blur due to rapid flight maneuvers, vibrations, and camera panning, which can significantly degrade downstream tasks such as target detection. Our goal is to explore a computationally-efficient and effective video deblurring approach to enhance UAV target detection performance.
To reduce computational cost, we first propose an \emph{Adaptive Latent Scale Selector} that dynamically adjusts the latent space resolution according to the intensity of UAV motion, thus balancing detail preservation with inference efficiency. To ensure temporal consistency, we introduce a \emph{Multi-Frame Alignment and Learnable Gating} module to warp and gate the preceding frames, allowing the model to fuse only relevant temporal information and suppress misaligned or uninformative features. Our method can effectively recover sharp details from the UAV video stream. Extensive experiments on real UAV benchmarks demonstrate that our method not only yields superior deblurring performance but also significantly boosts target detection accuracy, making it highly applicable to robust aerial vision tasks. Code will be publicly available \href{https://github.com/ZHIQIANGHU2021/Motion-Aware-Diffusion}{here}.

\end{abstract}
%%%%%%%%% BODY TEXT
\section{Introduction}
\label{sec:intro}

\begin{figure}[t]
	\centering
	\includegraphics[width=0.9\linewidth]{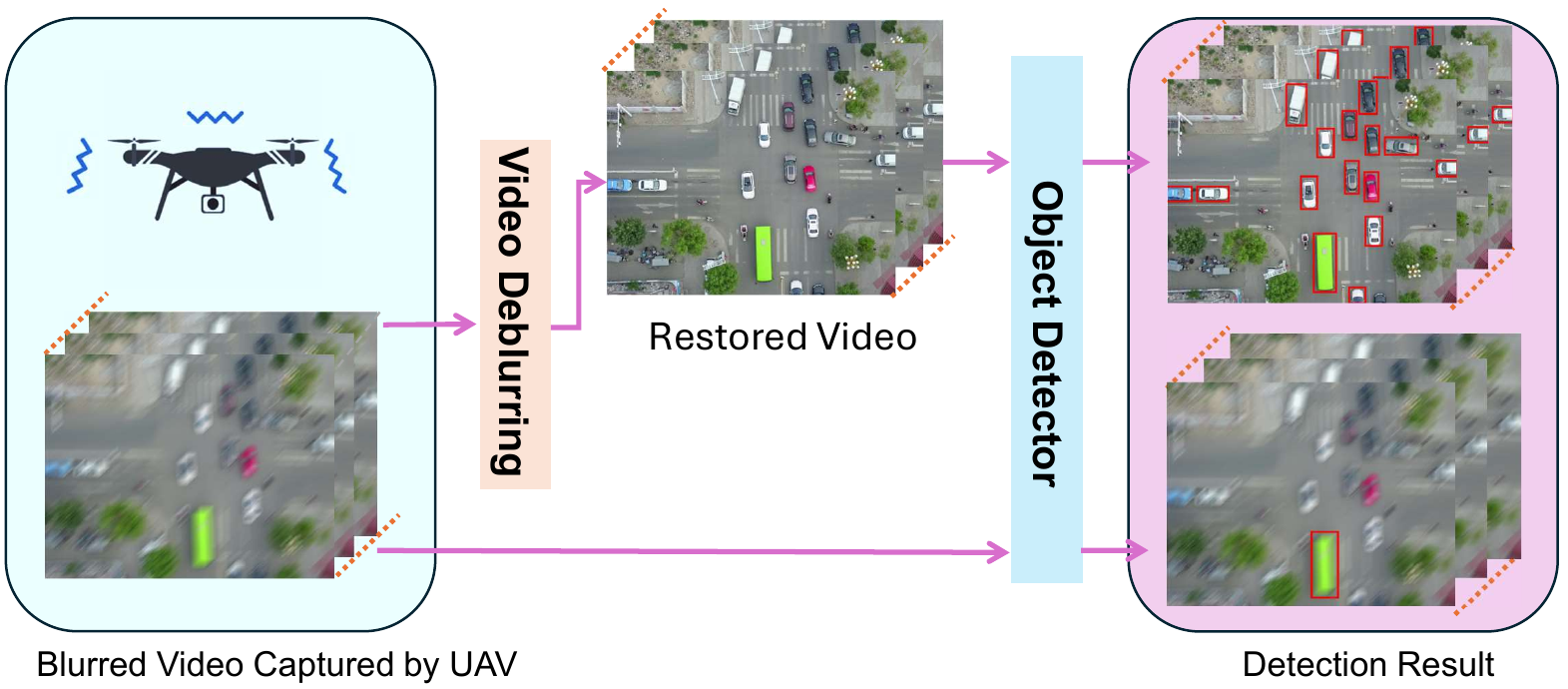}
	%\vspace{-4mm}
	\caption{\textbf{Impact of Video Deblurring on UAV-Based Object Detection.} 
    The left section shows a UAV capturing a blurry scene. The middle section represents the deblurring process, restoring sharp frames. The right section compares object detection results on both blurry and restored frames, demonstrating improved object detection performance with deblurring.}
	\label{fig:motivation}
 \vspace{-2mm}
\end{figure}

Unmanned Aerial Vehicles (UAVs) have become indispensable in various mission-critical fields, such as disaster surveillance, urban monitoring, and environmental inspection. In these applications, \emph{target detection} (e.g., identifying pedestrians, vehicles, or infrastructure anomalies) is vital for effective decision-making and timely interventions. However, UAV-captured videos often suffer from \textbf{severe motion blur} due to rapid flight maneuvers, vibrations, and dynamic environmental factors. Such motion blur not only degrades the visual quality but also poses a serious challenge to downstream detection algorithms, which rely on clear spatial details to localize and classify objects accurately.
Consequently, computationally efficient and effective video deblurring methods are highly valuable for enhancing UAV target detection applications as illustrated in Figure~\ref{fig:motivation}.

Traditional video deblurring techniques often prove inadequate in dynamic scenarios because they struggle to adapt to unpredictable environments and rapid UAV motion \cite{xu2010two}. In particular, they often find it challenging to optimize complex temporal models and generalize to the diverse types of motion blur encountered in real-world UAV scenarios.

\par To address these challenges, deep learning-based methods have emerged as a more powerful alternative. Approaches such as EDVR \cite{wang2019edvr} utilize advanced techniques like deformable convolution and gradual refinement schemes to tackle issues related to temporal alignment and motion compensation. Despite their success, these methods predominantly focus on the spatial domain and tend to overlook the potential of temporal information ~\cite{hu2022learning}. This oversight can limit their effectiveness, particularly in scenarios involving complex and unpredictable motion blur like UAV target detection and tracking.

Recently, Diffusion Models (DMs) have demonstrated remarkable performance in image synthesis~\cite{dhariwal2021diffusion} and restoration tasks~\cite{kawar2022denoising, chen2024hierarchical}. DMs generate high-fidelity images through a stochastic iterative denoising process, starting from pure Gaussian noise~\cite{dhariwal2021diffusion}. Compared to other generative models like  Generative Adversarial Networks (GANs), DMs offer a more accurate target distribution without the common issues of optimization instability or mode collapse~\cite{dhariwal2021diffusion}. However, DMs often face criticism for their high computational costs, as the iterative denoising steps become prohibitively expensive when dealing with high-resolution inputs~\cite{hoogeboom2023simple}. This limitation is especially problematic for UAV-based applications, where real-time or near real-time processing is crucial for timely decision-making in dynamic environments. Moreover, maintaining temporal coherence across consecutive frames remains challenging for DMs; they may produce outputs with inconsistent motion flow, manifesting as jitter or ghosting effects. In UAV scenarios involving rapid camera movements and frequent viewpoint changes, these artifacts severely compromise the reliability of subsequent tasks such as target detection or situational awareness.

\begin{figure}[t]
	\centering
	\includegraphics[width=1\linewidth]{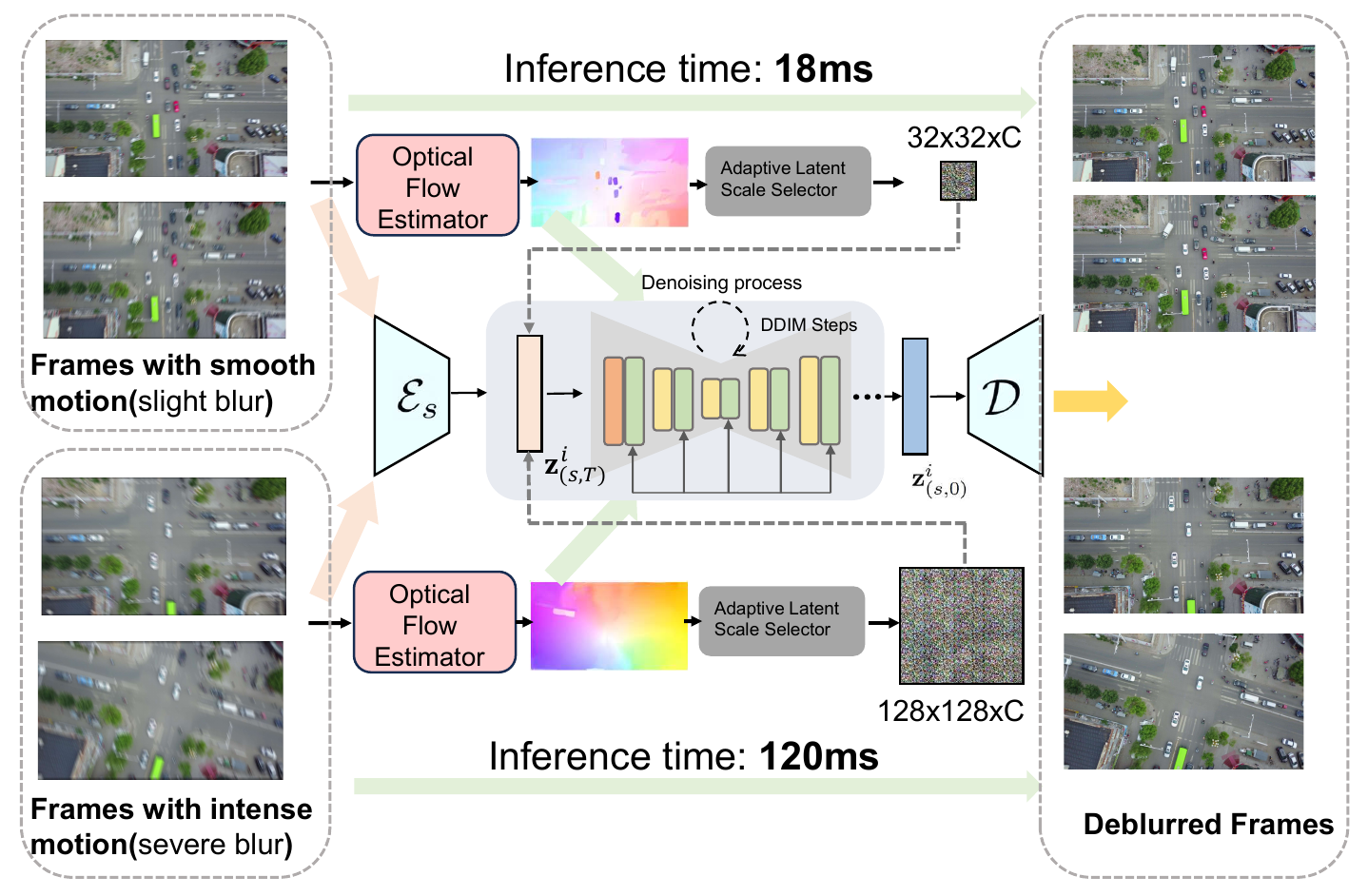}
	%\vspace{-4mm}
	\caption{\textbf{Inference Process with Adaptive Latent Feature Sizes:}  For frames with smooth motion (top), the Optical Flow Estimator calculates a lower motion intensity, leading the Adaptive Latent Scale Selector to choose a smaller latent feature size for encoding, which results in a shorter inference time; conversely, for frames with intense motion, a larger latent feature size is selected, leading to a longer inference time.}
	\label{fig:inference}
 \vspace{-2mm}
\end{figure}

To tackle these challenges, our work introduces a novel video deblurring framework aiming for UAV target detection tasks.
Firstly, to reduce computational complexity, we propose an \textbf{Adaptive Latent Scale Selector} module that dynamically adjusts the latent feature scale based on the intensity of motion detected from the optical flow in the blurry video frames, see Figure~\ref{fig:inference} for detailed information. This method allows our model to preserve fine details in regions of rapid motion while optimizing computational resources in more stable areas, thereby accelerating the inference process.
\par Secondly, to ensure temporal consistency across video frames, our method explicitly aligns preceding frames to the current view and employs a learnable gating mechanism to filter out misaligned or degraded cues. This approach recovers consistent temporal details and mitigates artifacts such as ghosting and jitter, ultimately enhancing video clarity and directly boosting target detection performance.  
Overall, the key contributions of our proposed method are:
\\ \textbf{Adaptive Latent Scale Selector (ALSS):} A motion-aware mechanism that dynamically adjusts the latent-space resolution based on motion intensity estimated from optical flow. ALSS strikes an effective balance between computational efficiency and detail preservation, please refer to Figure ~\ref{fig:inference} for the concept of our method.
\\ \textbf{Multi-Frame Alignment with Learnable Gating (MALG):} UAV video is prone to rapid viewpoint changes and dynamic backgrounds, making simple frame fusion insufficient. By adaptively filtering out degraded cues and aggregating multiple frames by Multi-Frame Cross-Attention, our method can recover temporally consistent details and mitigate artifacts.
\\ \textbf{Improved Detection Outcomes:} By effectively reducing severe blur while maintaining motion continuity, our approach substantially enhances detection performance on UAV video benchmarks. The recovered spatio-temporal fidelity not only produces sharper visual outputs but also translates into higher target detection accuracy.

%-------------------------------------------------------------------------
\section{Related Works}
\subsection{UAV Object Detection}

Recent advancements in UAV object detection have garnered significant attention due to their immense practical value and broad range of applications. \textbf{ViT-YOLO} \cite{zhang2021vit} introduces a Transformer-based approach for UAV object detection, effectively capturing global contextual information. However, its substantial computational cost poses significant challenges for real-time deployment on UAV platforms. To address this, \textbf{Drone-YOLO} \cite{zhang2023drone} optimizes feature extraction by integrating shallow features more effectively, thereby improving small object detection and inference speed. Similarly, \textbf{YOLOv8-s} \cite{li2023modified} enhances parameter efficiency by incorporating Bi-FPN and Ghost blocks, reducing computational overhead without compromising accuracy. Furthermore, \textbf{EFPN} \cite{deng2021extended} introduces a Feature Texture Transfer (FTT) module to refine shallow feature mapping, further boosting detection precision. Real-Time DEtection TRansformer (RT-DETR) ~\cite{zhao2024detrs} has emerged as a powerful end-to-end object detector, bridging the gap between efficiency and accuracy in real-time detection tasks. 
\\Despite these advancements, \textbf{none of these methods explicitly address the impact of motion blur on object detection performance}, which is particularly critical in UAV applications where rapid camera movement and environmental conditions can introduce significant blurring artifacts. Our work, by contrast, focuses on explicitly reducing blur to enhance detection, bridging this overlooked gap.

\subsection{Video Deblurring}
Video deblurring has been advanced significantly with deep learning, several approaches leveraged spatial and temporal coherence to improve performance over single-image techniques ~\cite{hu2022learning}. Early works by Su \textit{et al.}~\cite{su2017deep} focused on aligning and aggregating information across frames to enhance deblurring.

More recently, transformer-based architectures have emerged as a robust solution for video deblurring, image enhancement ~\cite{hu2024dynamic} capable of modeling long-range dependencies. Liang \textit{et al.}~\cite{liang2022recurrent} introduced a recurrent video restoration transformer with guided deformable attention, effectively leveraging both spatial and temporal information. Lin \textit{et al.}~\cite{lin2022flow} proposed a flow-guided sparse transformer that uses optical flows to guide the attention module, enhancing the model's ability to manage diverse motion blur conditions.
However, these methods often struggle to fully adapt to varying motion intensities across different regions of a video frame, limiting their real-world effectiveness. Our approach addresses this by adapting the processing scale based on motion intensity, preserving fine details in rapid-motion frames while conserving resources in stable frames.
%%%%%%%%%%%%%%%%

%%%%%%%%%%%%%%%%%%
\begin{figure*}[tbp]
	\centering
	\includegraphics[width=0.96\linewidth]{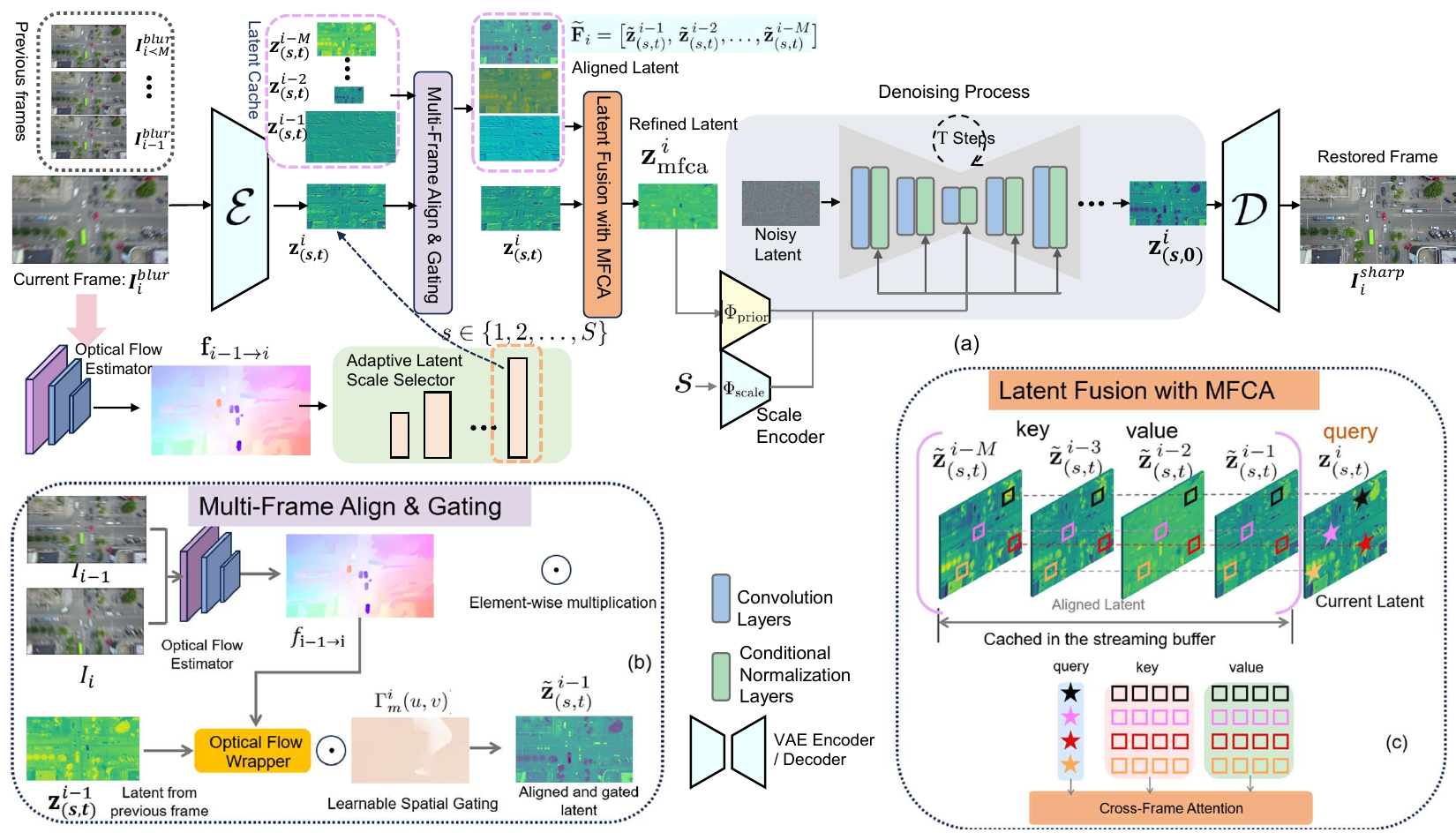}
	%\vspace{-4mm}
	\caption{ \textbf{Overview of Our Motion-Aware Dynamic Latent Diffusion Framework (a) for UAV Video Deblurring.}
		First, each blurry frame is encoded at multiple scales. Then, an \textbf{Adaptive Latent Scale Selector} (Sec.~III-C) dynamically determines the appropriate latent feature size based on the estimated motion intensity from optical flow, balancing fine detail preservation and computational efficiency. Next, our \textbf{Multi-Frame Alignment and Learnable Gating (b) and Latent Fusion with MFCA (c)} (Sec.~III-E) warps preceding frames and filters out misaligned or uninformative regions, ensuring robust temporal fusion. Finally, we perform \textbf{diffusion-based denoising} (Sec.~III-E) conditioned on these fused latent and decode them back to the spatial domain to obtain deblurred frames. }
	\label{fig:overall_framework}
\end{figure*}
\subsection{Diffusion Models for Low-Level Vision}

Diffusion models (DMs) have emerged as a powerful tool in image restoration tasks, including super-resolution~\cite{li2022srdiff}, inpainting ~\cite{lugmayr2022repaint}, and deblurring~\cite{whang2022deblurring}, due to their capability to generate high-fidelity images through an iterative denoising process~\cite{dhariwal2021diffusion,song2021scorebased}. These models refine noisy inputs by reversing a diffusion process, which has been highly effective in various low-level vision tasks, as demonstrated by models like DDRM~\cite{kawar2022denoising}.
However, the application of DMs to high-resolution and video restoration tasks is hindered by their significant computational cost and the challenge of maintaining temporal consistency across frames.
Our approach tackles these challenges by preserving temporal coherence, reducing artifacts, and maintaining high-quality deblurring across frames.
%%%%%%%%%%%%%%%

%%%%%%%%%%%%%%% ICRA2025 METHOD (MODIFIED) START %%%%%%%%%%%%%%%
\section{Method}
\label{sec:method}
\subsection{Method Overview}
In UAV-based scenarios, formally, we have a blurry video sequence 
$\mathcal{I}^{\text{blur}} = \{\mathbf{I}^{\text{blur}}_1, \mathbf{I}^{\text{blur}}_2, \dots, \mathbf{I}^{\text{blur}}_N\}$,
where each frame $\mathbf{I}^{\text{blur}}_i \in \mathbb{R}^{H \times W \times C}$, \( H \), \( W \), and \( C \) denote the height, width, and number of color channels of the image, respectively. Our goal is to produce a deblurred sequence 
$\mathcal{I}^{\text{clear}} = \{\mathbf{I}^{\text{clear}}_1, \mathbf{I}^{\text{clear}}_2, \dots, \mathbf{I}^{\text{clear}}_N\}$,
which preserves spatial details and maintains temporal consistency for downstream UAV applications. 

\subsection{Adaptive Latent Scale Selector}
\subsubsection{Latent Extraction}
We encode each frame \(\mathbf{I}_i\) at scale \(s\) via 
$\mathbf{z}^{i}_{(s,t)} = \mathcal{E}_s(\mathbf{I}_i)$, 
where $s \in \{1,\dots,S\}$ is the chosen latent scale and $i \in \{1,\dots,N\}$ indicates frame index and \( t \) is the diffusion step.
Here, \(\mathcal{E}_s\) is the encoder that produces a latent representation at resolution scale \(s\). We adopt the auto-encoder architecture from Stable Diffusion~\cite{rombach2022high} to obtain multi-level latent features.

\subsubsection{Adaptive Latent Scale Selector}
\label{sec:adaptive_scale_selector}
To handle varying motion intensities across UAV video frames, we adaptively choose the most suitable latent resolution scale for each frame based on optical flow. Specifically, we estimate the forward optical flow 
$\mathbf{f}_{i-1 \to i} \in \mathbb{R}^{H \times W \times D_f}$ where $\mathbf{f}_{i-1 \to i}$ represents the flow vectors from frame $(i-1)$ to frame $i$, and $D_f$ is the dimensionality of the flow field. 
We use RAFT~\cite{teed2020raft} to compute $\mathbf{f}_{i-1 \to i}$ between two consecutive blurry UAV frames $\mathbf{I}^{\text{blur}}_{i-1}$ and $\mathbf{I}^{\text{blur}}_i$. 
\paragraph{Motion Intensity Estimation.} 
We define the motion intensity 
$\|\mathrm{Flow}(\mathbf{I}^{\text{blur}}_{i-1}, \mathbf{I}^{\text{blur}}_{i})\|$
as the magnitude of the optical flow vector:
\begin{equation}
\|\mathrm{Flow}(\mathbf{I}^{\text{blur}}_{i-1}, \mathbf{I}^{\text{blur}}_{i})\|
= \sqrt{(\Delta x)^2 + (\Delta y)^2},
\end{equation}
where $(\Delta x, \Delta y)$ are the  horizontal and vertical flow components from $\mathbf{f}_{i-1 \to i}$, respectively.
\paragraph{Scale Selection.}
Rather than relying on a fixed threshold, we map this motion intensity to a suitable latent scale $s_i$ by linearly interpolating within a predefined set of scales $\{1, \dots, S\}$. Specifically,
\begin{equation}
\label{eq:motion_scale_select}
s_i
=
\mathrm{round}\Bigl(
\tfrac{\|\mathrm{Flow}(\mathbf{I}^{\text{blur}}_{i-1}, \mathbf{I}^{\text{blur}}_{i})\|}
{\text{MaxFlow}_{\mathrm{adaptive}}(i)} \times (S-1)
\Bigr)
+ 1,
\end{equation}
where \( \mathrm{round}(\cdot) \) denotes rounding to the nearest integer. A larger flow magnitude indicates stronger motion and thus selects a larger scale $s$. 
We define $\text{MaxFlow}_{\mathrm{adaptive}}(i)$ as follows:
\begin{equation}
\label{eq:maxflow_adaptive}
\begin{aligned}
\text{MaxFlow}_{\mathrm{adaptive}}(i)
&= \alpha \cdot 
\Bigl\|\mathrm{Flow}\bigl(\mathbf{I}^{\text{blur}}_{i-1}, \mathbf{I}^{\text{blur}}_{i}\bigr)\Bigr\|_{\text{max}} \\
&\quad + (1 - \alpha)\,\text{MaxFlow}_{\mathrm{adaptive}}(i-1).
\end{aligned}
\end{equation}
where $\alpha$ is a smoothing factor $(0 < \alpha < 1)$ that balances current and previous flow magnitudes, and $\text{MaxFlow}_{\mathrm{adaptive}}(i-1)$ is the adaptive reference from the previous frame. The $\alpha$ is set to be 0.8 in our experiments. The selected scale $s_i$ thus adapts in real-time according to the UAV’s motion intensity, preserving fine details when movement is large while avoiding unnecessary high-resolution encodings during smooth flight segments.
\subsection{Latent Diffusion with Motion-Aware Conditioning}
\subsubsection{Multi-Frame Alignment with Learnable Gating}
\label{sec:multi_frame_gating}
In high-dynamic UAV scenarios, significant motion variations and complex environmental factors often lead to severe misalignment and occlusion across frames. To maintain consistency in the current blurred frame while leveraging clear region information from preceding frames, we propose a multiple-frame aggregation module. However, naively aggregating multiple frames without explicit handling of such discrepancies can result in substantial artifacts and temporal inconsistencies. To mitigate these challenges, we first spatially align each preceding frame to the current frame using optical flow-based warping. Additionally, we introduce a learnable spatial gating mechanism that adaptively suppresses unreliable regions while preserving informative, well-aligned features. This per-pixel gating strategy ensures that only temporally coherent and structurally consistent regions contribute to the final reconstruction, thereby enhancing robustness against motion-induced distortions and occlusions.

We take the previous \(M\) latent \(\{\mathbf{z}^{\,i-m}_{(s,t)} \mid m=1,\dots,M\}\). If any latent is at a different scale \(s' \neq s\), we first resize it to the target scale \(s\). Then, using the optical flow \(f_{(i-m)\to i}\), we warp the latent:
\begin{equation}
\hat{\mathbf{z}}_{(s,t)}^{\,i-m} 
= \mathcal{W}\!\Bigl(\mathrm{Resize}(\mathbf{z}^{\,i-m}_{(s',t)}),\, f_{(i-m)\to i}\Bigr).
\end{equation}
A three-layer convolutional network (with kernel sizes \(7\times7\), \(5\times5\), and \(3\times3\)) is used as the warping operator \(\mathcal{W}\). Next, a learnable mask \(\Gamma_m^i \in \mathbb{R}^{H\times W}\) is applied to obtain the gated latent:
\begin{equation}
\tilde{\mathbf{z}}_{(s,t)}^{\,i-m}(u,v) 
= \sigma\bigl(\Gamma_m^i(u,v)\bigr)\,\odot\,\hat{\mathbf{z}}_{(s,t)}^{\,i-m}(u,v),
\end{equation}
where \(\sigma(\cdot)\) is the sigmoid function and \(\odot\) denotes element-wise multiplication. We use a lightweight gating network to predict \( \Gamma_m^i \). We then stack these gated latent for following Multi-Frame Cross-Attention calculation:
\begin{equation}
\widetilde{\mathbf{F}}_i 
= \bigl[\tilde{\mathbf{z}}_{(s,t)}^{\,i-1},\, \tilde{\mathbf{z}}_{(s,t)}^{\,i-2}, \dots, \tilde{\mathbf{z}}_{(s,t)}^{\,i-M}\bigr].
\end{equation}
%%%%%%%%%%%%%%%%%
\subsubsection{Latent Fusion with Multi-Frame Cross-Attention (MFCA)}
We fuse information from past frames via Multi-Frame Cross-Attention (MFCA) see Figure 3(c). Specifically, we define $\mathbf{Q} = W^Q \,\mathrm{Flatten}\bigl(\mathbf{z}^{\,i}_{(s,t)}\bigr)$, 
$\mathbf{K} = W^K \,\mathrm{Flatten}\bigl(\widetilde{\mathbf{F}}_i\bigr)$, $\mathbf{V} = W^V \,\mathrm{Flatten}\bigl(\widetilde{\mathbf{F}}_i\bigr)$,
where \(W^Q, W^K, W^V\) map each flattened input into a common \(d\)-dimensional space. The MFCA operation is then computed as
\[
\mathrm{MFCA}(\mathbf{Q},\,\mathbf{K},\,\mathbf{V})
= \mathrm{softmax}\Bigl(\tfrac{\mathbf{Q}\mathbf{K}^\top}{\sqrt{d}}\Bigr)\,\mathbf{V},
\]
and we reshape the output back to a 2D grid to obtain the refined latent \(\mathbf{z}^{\,i}_{\mathrm{mfca}}\).
%%%%%%%%%%%%%%%%%
\subsubsection{Scale-Conditioned Reverse Denoising}
To enable the denoising network (UNet) to explicitly handle different spatial scales, we introduce a \emph{scale embedding} \(e_s\). Specifically, we learn a small mapping \(\Phi_{\mathrm{scale}}(s)\) that converts the discrete scale index \(s\) into a feature vector \(e_s\). This embedding is injected into the UNet via conditional normalization layers; for instance, at an intermediate layer with feature map \(F\), the scale-conditioned feature is computed as
\begin{equation}
F' = \gamma(e_s)\odot F + \beta(e_s),
\end{equation}
where \(\gamma(e_s)\) and \(\beta(e_s)\) are generated by small networks conditioned on \(e_s\). Furthermore, in addition to the scale embedding, we also inject the \emph{refined latent} as a conditional signal. Similar to scale conditioning, we feed the \(\mathbf{z}^{\,i}_{\mathrm{mfca}}\) into a small MLP denoted as 
\(\Phi_{\mathrm{prior}}(z)\) to produce adaptive parameters \(\gamma_c\) and \(\beta_c\). Then, at another intermediate layer (or the same layer), the feature map can be updated by
\begin{equation}
F'' = \gamma_c(\mathbf{z}^{\,i}_{\mathrm{mfca}}) \odot F' + \beta_c(\mathbf{z}^{\,i}_{\mathrm{mfca}}),
\end{equation}
where \(F'\) is the feature already modulated by \(e_s\). This helps preserve structural information from the original input, providing an additional prior for better deblurring.
Finally, this refined latent is fed into our UNet-like denoiser along with the diffusion timestep \(t\), the scale embedding \(e_s\), and optionally the clean latent \(\mathbf{z}_{(s,0)}\). Formally, the denoiser is defined as:
\begin{equation}
\label{eq:denoiser_cond}
\epsilon_\theta\bigl(\mathbf{z}^{\,i}_{\mathrm{mfca}},\,t,\,e_s,\,\mathbf{z}_{(s,0)}\bigr) 
= \mathrm{UNet}\!\Bigl(\mathbf{z}^{\,i}_{\mathrm{mfca}},\,t,\,e_s,\,\mathbf{z}_{(s,0)}\Bigr),
\end{equation}
where the UNet incorporates both the scale information and the refined latent via conditional normalization. With the predicted noise, we perform the standard reverse diffusion update (as in DDPM~\cite{ho2020denoising}):
\begin{equation}
\begin{split}
\mathbf{z}^{\,i}_{(s,t-1)} &= \frac{1}{\sqrt{\alpha_t}} \Bigl[\mathbf{z}^{\,i}_{(s,t)} - \frac{1-\alpha_t}{\sqrt{1-\bar{\alpha}_t}}\,\epsilon_\theta\bigl(\mathbf{z}^{\,i}_{\mathrm{mfca}},\,t,\,e_s,\,\mathbf{z}_{(s,0)}\bigr)\Bigr] \\
&\quad + \sigma_t\,\eta,
\end{split}
\end{equation}
where \(\eta\sim\mathcal{N}(0,I)\), \(\bar{\alpha}_t=\prod_{\tau=1}^{t}\alpha_\tau\), and \(T\) is the total number of reverse diffusion steps. After iterating down to \(t=0\), we obtain the deblurred latent \(\mathbf{z}^{\,i}_{(s,0)}\) then decode it via \(\mathcal{D}\) to yield the deblurred frame \(I_i^{\mathrm{clear}}\).
%%%%%%%%%%%%%%%%%%%%%%%%%%%%%%%%%%%%

\subsection{Training Strategy}
\label{sec:training}

\noindent
\textbf{Phase 1: Pre-training Encoder and Decoder.} 
We first train an auto-encoder $(\mathcal{E}, \mathcal{D})$ on ground-truth data by minimizing 
$
\mathcal{L}_{\mathrm{rec}} 
= \Bigl\|\mathcal{D}\bigl(\mathcal{E}(I^{\mathrm{clear}})\bigr) - I^{\mathrm{clear}}\Bigr\|_2^2,
$
ensuring that the reconstructed output closely matches the ground-truth frame. This phase yields a high-quality VAE capable of encoding images into latent $\mathbf{z}_{(s,0)}$ and decoding them back.

\noindent
\textbf{Phase 2: Multi-Scale Diffusion Denoising Model Training.} 
To handle varying latent resolutions, we adopt a multi-scale training scheme. Specifically, we randomly sample a scale $s$ from the discrete set $\{1,\ldots,S\}$, then encode the blurry input frame $I_i^{\mathrm{blur}}$ using $\mathcal{E}_s$ to obtain 
$
\mathbf{z}_{(s,0)}^i = \mathcal{E}_s\bigl(I_i^{\mathrm{blur}}\bigr).
$
We perform forward diffusion over $T$ steps to corrupt $\mathbf{z}_{(s,0)}^i$ into $\mathbf{z}_{(s,t)}^i$ by adding noise. During training, we compute the fused latent $\mathbf{z}_{(s,t)}^i \!\rightarrow\! \mathbf{z}_{\mathrm{mfca}}^i$ (via Multi-Frame Cross-Attention, see Sec.~\ref{sec:multi_frame_gating}) and inject both the scale embedding $e_s$ and the refined latent (optionally combined with the clean latent $\mathbf{z}_{(s,0)}^i$) into the denoiser $\epsilon_\theta$. The objective is to predict the added noise:
$
\mathcal{L}_{\mathrm{denoise}}
= \mathbb{E}_{s \sim p(s),\,t,\,\epsilon}
\Bigl\|\epsilon 
- \epsilon_\theta\bigl(\mathbf{z}_{\mathrm{mfca}}^i,\,t,\,e_s,\mathbf{z}_{(s,0)}^i\bigr)\Bigr\|_2^2,
$
where $p(s)$ is typically uniform (or can be biased to favor certain scales). This multi-scale training strategy enables the diffusion UNet $\epsilon_\theta(\cdot)$ to handle diverse resolutions and leverage multi-frame temporal cues.

\noindent
\textbf{Phase 3: Joint Fine-Tuning.} 
After obtaining a preliminary denoising model, we fine-tune \emph{all} modules, including the set of encoders $\{\mathcal{E}_s\}$ for each scale, the decoder $\mathcal{D}$, and the diffusion UNet $U_\theta$ under a combined objective:$
\mathcal{L}_{\mathrm{total}} 
= \lambda_1\,\mathcal{L}_{\mathrm{rec}} 
+ \lambda_2\,\mathcal{L}_{\mathrm{denoise}},
$where $\lambda_1$ and $\lambda_2$ are weights balancing VAE reconstruction fidelity and noise-prediction accuracy, respectively. We set $\lambda_1=0.6$ and $\lambda_2=0.4$ throughout our experiments. This joint fine-tuning ensures that the latent space (VAE) and the scale-conditioned diffusion model are well-aligned, leading to robust UAV video deblurring across different motion intensities and resolutions.

%%%%%%%%%%%%%%% ICRA2025 METHOD (MODIFIED) END %%%%%%%%%%%%%%%

\begin{figure}[tbp]
	\centering
	\includegraphics[width=1\linewidth]{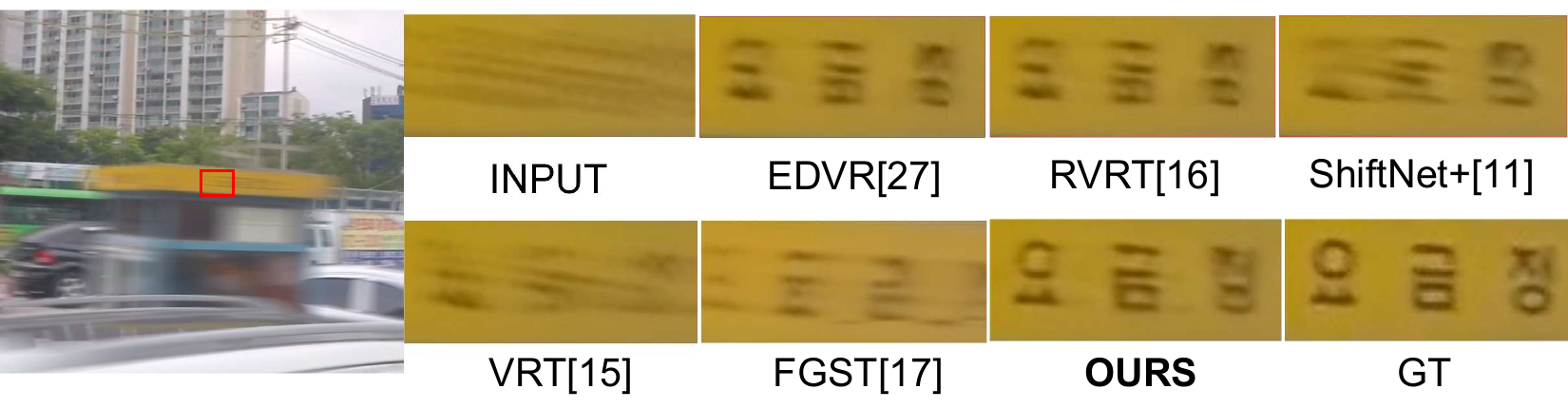}
	\vspace{-4mm}
	\caption{\textbf{Visual Comparisons Between our Method and SOTA Methods on GoPro ~\cite{nah2017deep} Dataset.} 
}
	\label{fig:compare_SOTA_GoPro}
\end{figure}

%%%%%%%%%%%%
\begin{figure}[tbp]
	\centering
	\includegraphics[width=1\linewidth]{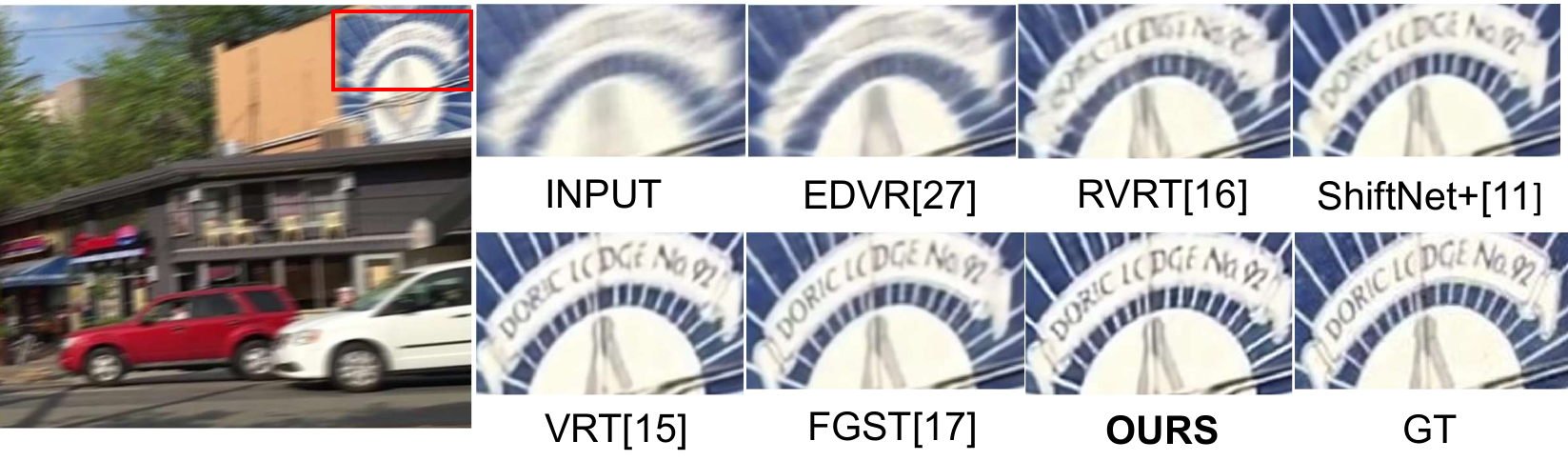}
	\vspace{-4mm}
	\caption{\textbf{Visual Comparisons Between Our Method and SOTA Methods on DVD ~\cite{su2017deep} Dataset.} 
}
	\label{fig:compare_SOTA_DVD}
 \vspace{-6mm}
\end{figure}

\begin{figure}[htbp]
	\centering
	\includegraphics[width=1\linewidth]{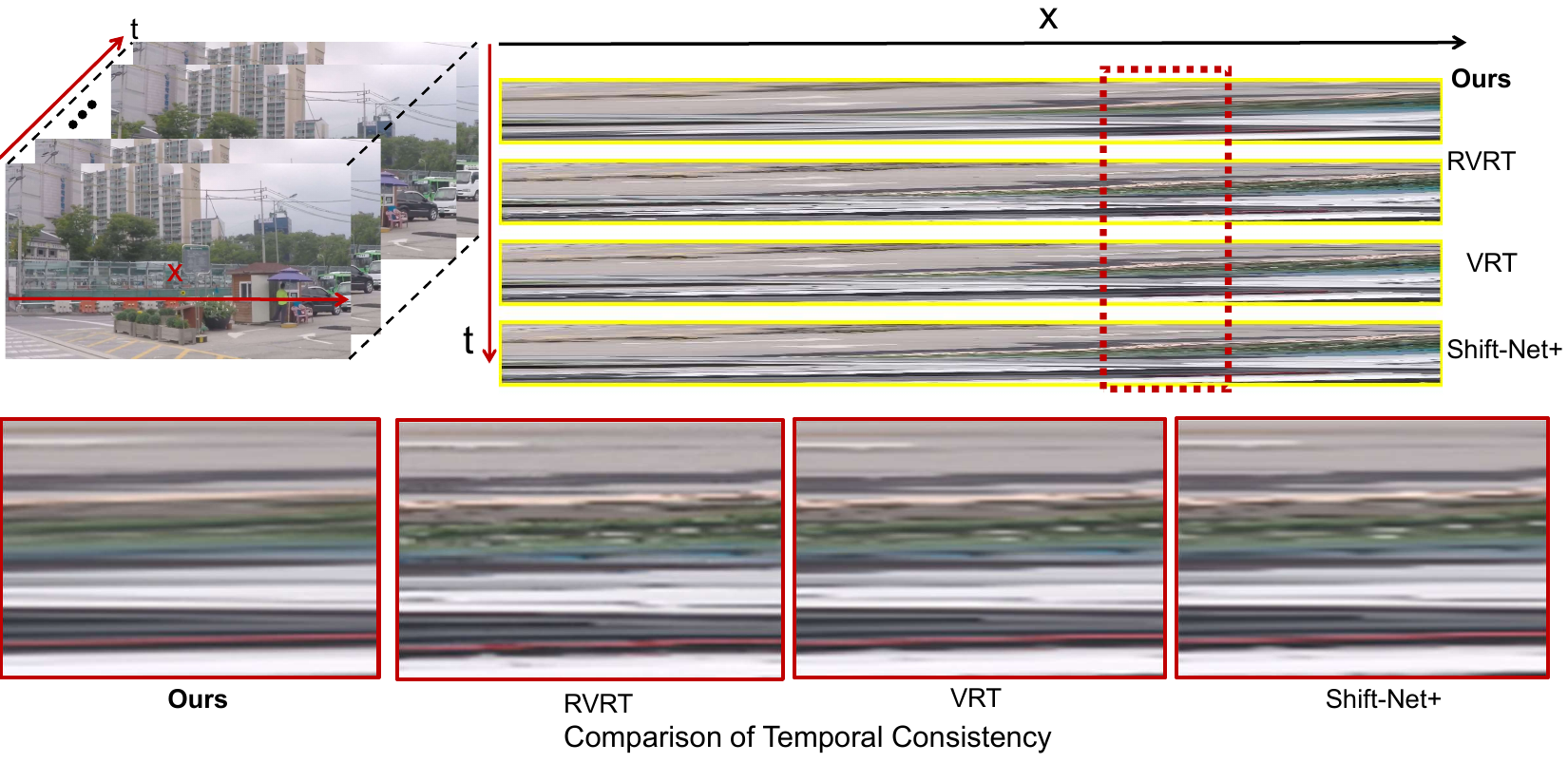}
	%\vspace{-4mm}
	\caption{\textbf{Temporal Consistency Comparison on DVD ~\cite{su2017deep} dataset.} We analyze the temporal profile along a fixed row in consecutive frames to assess temporal consistency. The temporal profiles generated by existing methods, including \textbf{VRT}~\cite{liang2024vrt}, \textbf{RVRT}~\cite{liang2022recurrent}, and \textbf{Shift-Net+}~\cite{li2023simple}, reveal noticeable distortions and discontinuities, indicating flickering artifacts across frames. In contrast, \textbf{Ours} maintains a smooth and coherent profile over time, showcasing improved temporal consistency. }
	\label{fig:Consistency}
\end{figure}

\begin{table}[htbp]
  \caption{\textbf{Quantitative comparisons on the GoPro and DVD datasets.}
  The best results are highlighted in bold.}
  \vspace{-2 mm}
  \centering
  \scalebox{0.9}{
  \begin{tabular}{|l|cc|cc|}
    \toprule
    \multirow{2}{*}{Method}
    & \multicolumn{2}{c|}{GoPro}
    & \multicolumn{2}{c|}{DVD} \\
    \cmidrule(lr){2-3} \cmidrule(lr){4-5}
    & PSNR & SSIM
    & PSNR & SSIM \\
    \midrule
    STFAN~\cite{zhou2019spatio}
        & 28.69 & 0.8610 & 31.24 & 0.9340 \\
    STDA~\cite{zhang2022spatio}
        & 32.62 & 0.9375 & 33.05 & 0.9374 \\
    NAFNet~\cite{chen2022simple}
        & 33.69 & 0.9670 & -     & -      \\
    ARVo~\cite{li2021arvo}
        & -     & -      & 32.80 & 0.9352 \\
    VRT~\cite{liang2024vrt}
        & 34.81 & 0.9724 & 34.27 & 0.9651 \\
    RVRT~\cite{liang2022recurrent}
        & 34.92 & 0.9738 & 34.30 & 0.9655 \\
    Shift-Net+~\cite{li2023simple}
        & 35.88 & 0.9790 & 34.69 & 0.9690 \\
    \textbf{Ours}
        & \textbf{36.12} & \textbf{0.9799}
        & \textbf{35.18} & \textbf{0.9778} \\
    \bottomrule
  \end{tabular}
  }
  \label{tab:combined_gopro_dvd}
  \vspace{-2 mm}
\end{table}

%%%%%%%%%%%%%%%ICRA2025 METHOD END%%%%%%%%%%%%%%%%
\begin{figure*}[tbp]
	\centering
	\includegraphics[width=0.8\linewidth]{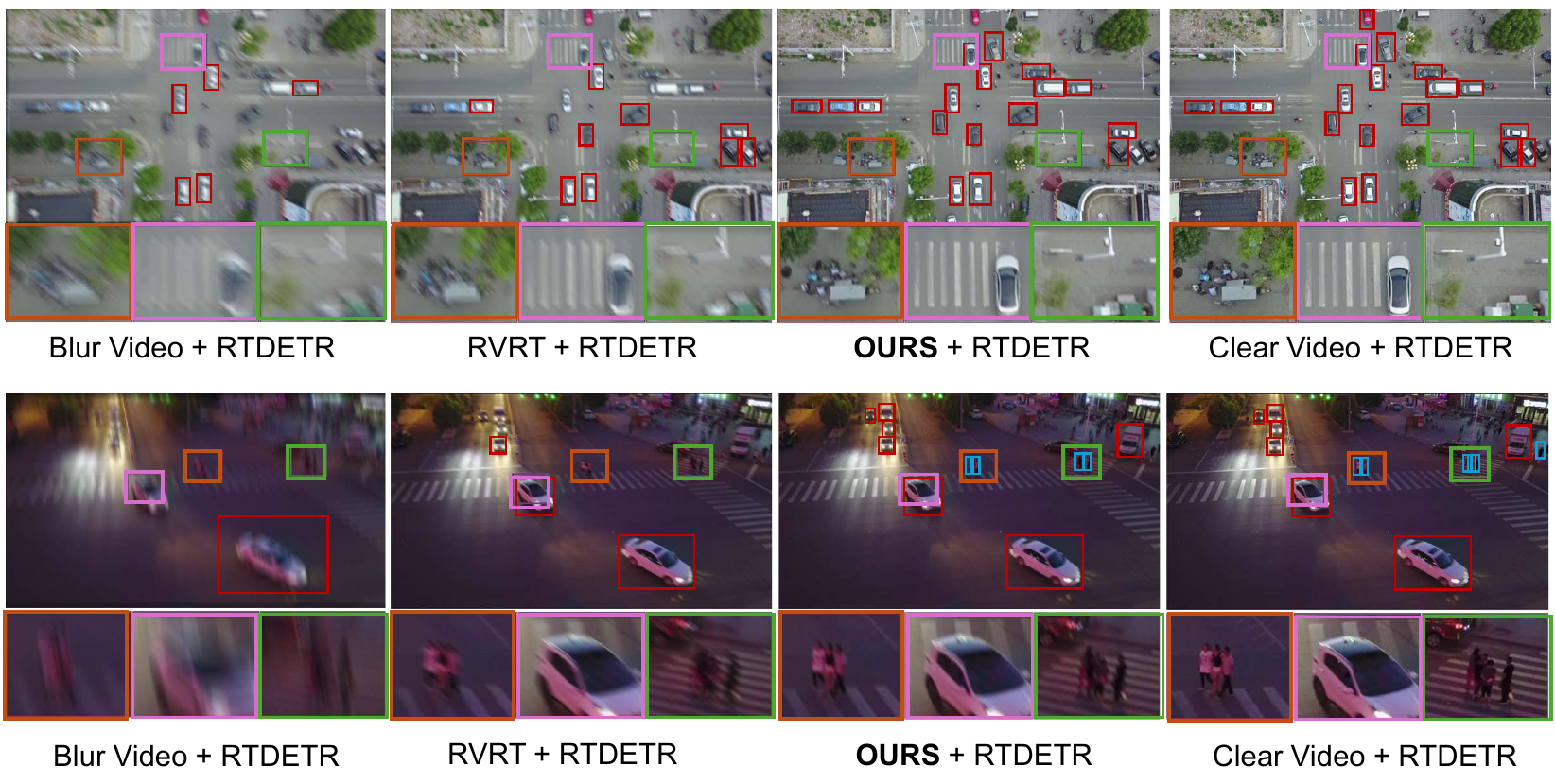}
	%\vspace{-4mm}
	\caption{\textbf{Object Detection Performance on the \textbf{VisDrone-VID2019} Dataset~\cite{zhu2021detection}}. 
        From left to right: (1) Detection results on blurred input frames using RTDETR ~\cite{zhao2024detrs}, (2) RVRT~\cite{liang2022recurrent} deblurred frames + RTDETR detection ~\cite{zhao2024detrs}, (3) \textbf{Our method} deblurred frames + RTDETR detection, and (4) Detection results on original clear frames using RTDETR ~\cite{zhao2024detrs}.
        Our approach (\textbf{Ours + RTDETR ~\cite{zhao2024detrs}}) achieves the best detection performance, effectively recovering motion-degraded objects while preserving fine details. 
        The zoomed-in patches further highlight that our deblurring algorithm restores sharper textures and more accurate object structures compared to RVRT~\cite{liang2022recurrent}.
}
	\label{fig:VisDrone}
 \vspace{-1mm}
\end{figure*}
%%%%%%%%%%%%%%%

\section{Experiments}

\subsection{Datasets}

\noindent \textbf{GoPro Dataset:} The GoPro dataset~\cite{nah2017deep} contains 3,214 pairs of blurry and sharp images with a resolution of 1280 \(\times\) 720, split into 2,103 training pairs and 1,111 testing pairs. It is a standard benchmark for video deblurring, capturing real-world dynamic scenes with significant motion blur.

\noindent \textbf{DVD Dataset:} The DVD dataset~\cite{su2017deep} includes 71 videos, yielding 6,708 blurry-sharp frame pairs, with 5,708 pairs for training and 1,000 pairs for testing. It covers a wide range of motion patterns and blur levels, providing a comprehensive evaluation environment for deblurring algorithms.

\noindent \textbf{VisDrone-VID2019 Dataset:}
We also conduct experiments on the \textbf{VisDrone-VID2019}~\cite{zhu2021detection} benchmark, which comprises 79 real-world UAV video sequences, totaling 33,366 frames. These videos are captured under diverse weather and illumination settings across multiple urban locations, ensuring a broad range of motion patterns and scene complexity. 

To simulate motion blur for our method’s evaluation, we apply the approach of \cite{brooks2019learning} to generate artificially blurred frames from the original VisDrone-VID2019 videos. This procedure enables us to assess how effectively our deblurring framework can restore image clarity in high-speed UAV footage and subsequently improve object detection. For quantitative assessment, we follow the protocol of the MS COCO-style metrics—namely \textbf{AP}, \textbf{AP50}, \textbf{AP75}, \textbf{AR1}, \textbf{AR10}, \textbf{AR100}, and \textbf{AR500} where AP (averaged over IoU thresholds from 0.5 to 0.95) serves as the primary ranking metric. 

\subsection{Implementation Details}

\noindent \textbf{Training Details:}
The network is implemented in PyTorch and trained on 8 NVIDIA A100 GPUs with a batch size of 8. The initial learning rate is set to $4 \times 10^{-4}$. We use the Adam optimizer~\cite{kingma2014adam} with $\beta_1 = 0.9$ and $\beta_2 = 0.999$. The flow estimator in our method utilizes pre-trained weights from RAFT~\cite{teed2020raft} and remains fixed during training. During training, input images are randomly cropped to $256 \times 256$ patches with random flipping and rotation augmentations.
\begin{figure}[htbp]
	\centering
	\includegraphics[width=0.8\linewidth]{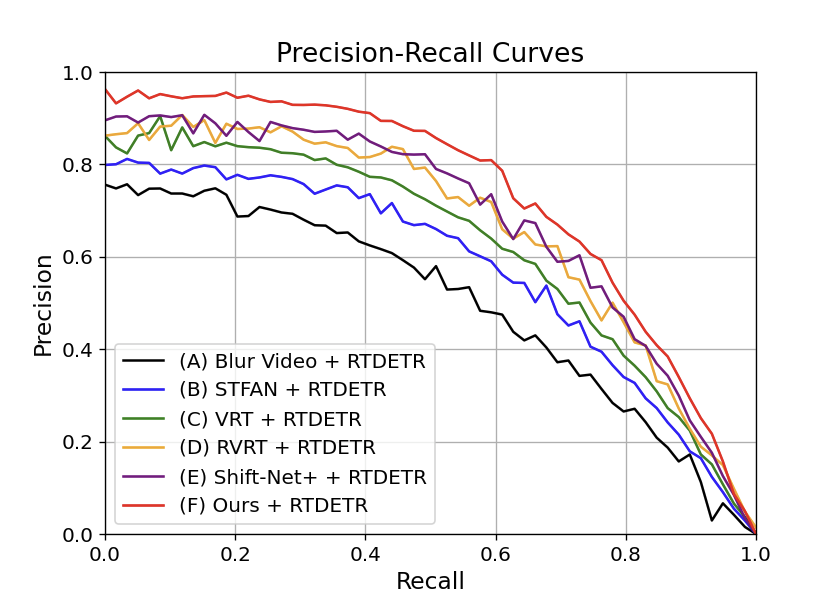}
	\vspace{-4mm}
	\caption{\textbf{Precision-recall Curves on the \textbf{VisDrone-VID2019 ~\cite{zhu2021detection}} Dataset.} The proposed method (\textbf{F: Ours + RTDETR ~\cite{zhao2024detrs}}) achieves the best precision-recall performance, consistently outperforming previous methods including STFAN ~\cite{zhou2019spatio}, VRT ~\cite{liang2024vrt}, RVRT ~\cite{liang2022recurrent}, and Shift-Net~\cite{li2023simple}. This demonstrates its superior capability in handling degraded UAV imagery and improving object detection accuracy.
}
	\label{fig:ROC}
 \vspace{-3mm}
\end{figure}
\begin{table*}[htbp]
\centering
\caption{
\textbf{Comparison results (AP, AR \%) on the VisDrone-VID2019~\cite{zhu2021detection} dataset.} 
All entries use the same RTDETR ~\cite{zhao2024detrs} detection framework. 
We report AP, AP$_{50}$, AP$_{75}$, and AR under different candidate limits (1, 10, 100, 500).
}
\label{tab:visdrone_rtdetr}
\setlength{\tabcolsep}{3pt}
\small{
\begin{tabular}{|l|cccc|ccc|}
\hline
\multirow{2}{*}{Method} & 
\multirow{2}{*}{AP} & 
\multirow{2}{*}{AP$_{50}$} & 
\multirow{2}{*}{AP$_{75}$} & 
\multirow{2}{*}{AR$_{1}$} & 
\multirow{2}{*}{AR$_{10}$} & 
\multirow{2}{*}{AR$_{100}$} & 
\multirow{2}{*}{AR$_{500}$} \\
& & & & & & & \\
\hline
(A) Blur Video + RTDETR ~\cite{zhao2024detrs} 
& 21.53 & 45.22 & 16.34 & 10.45 & 28.74 & 42.66 & 44.12 \\
(B) STFAN~\cite{zhou2019spatio} + RTDETR ~\cite{zhao2024detrs} 
& 29.22 & 58.00 & 25.34 & 14.30 & 35.58 & 50.75 & 53.67 \\
(C) VRT~\cite{liang2024vrt} + RTDETR ~\cite{zhao2024detrs} 
& 31.50 & 60.20 & 27.80 & 15.20 & 37.44 & 52.20 & 55.10 \\
(D) RVRT~\cite{liang2022recurrent} + RTDETR ~\cite{zhao2024detrs} 
& 32.60 & 62.80 & 28.60 & 16.10 & 38.05 & 53.60 & 56.00 \\
(E) Shift-Net+~\cite{li2023simple} + RTDETR ~\cite{zhao2024detrs} 
& 33.80 & 63.20 & 29.25 & 16.60 & 38.90 & 54.40 & 57.50 \\
(F) \textbf{Ours} + RTDETR ~\cite{zhao2024detrs} 
& \textbf{35.12} & \textbf{64.70} & \textbf{30.80} & \textbf{17.75} & \textbf{40.10} & \textbf{55.90} & \textbf{59.20} \\
\hline
\end{tabular}
}
\vspace{-2mm}
\end{table*}

\subsection{Main Results}

\subsubsection{Quantitative Evaluation}

\noindent \textbf{GoPro Dataset:}
As reported in Table~\ref{tab:combined_gopro_dvd}, our proposed method achieves a PSNR/SSIM of \textbf{36.12}~dB and \textbf{0.9799}, outperforming existing state-of-the-art approaches such as Shift-Net+~\cite{li2023simple} and RVRT~\cite{liang2022recurrent}. Notably, our framework surpasses these competitors by a clear margin (e.g., a relative PSNR gain of over 0.2\,dB compared to Shift-Net+), indicating its superior capacity for recovering details under diverse motion blur patterns. Visual comparisons in Figure~\ref{fig:compare_SOTA_GoPro} further highlight the sharper textures and reduced artifact levels of our outputs.

\noindent \textbf{DVD Dataset:}
As shown in Table~\ref{tab:combined_gopro_dvd}, our approach also maintains a clear advantage over prior works on the DVD dataset, achieving \textbf{35.18}~dB in PSNR and \textbf{0.9778} in SSIM. Compared to alternative methods, our framework consistently restores finer structural details and suppresses temporal artifacts even in challenging cases with large camera motion and dynamic objects.

%%%%%%%%%%%%%%%%%%%%%%%%%
\begin{table*}[htbp]
\centering
\caption{\textbf{Ablation Study on the GoPro Dataset (PSNR/SSIM) and the VisDrone-VID2019 Dataset.} 
ALSS = Adaptive Latent Scale Selector, 
MAlign = Multi-Frame Alignment, 
Gating = Learnable Gating. 
The best results are in \textbf{bold}.}
\label{tab:ablation_combined}
\setlength{\tabcolsep}{3pt} % 调小列间距
\scalebox{0.9}{ % 整体缩放
\begin{tabular}{|l|c|c|c|cc|ccccccc|}
\hline
\multirow{2}{*}{Method} 
& \multirow{2}{*}{ALSS} 
& \multirow{2}{*}{MAlign} 
& \multirow{2}{*}{Gating} 
& \multicolumn{2}{c|}{GoPro} 
& \multicolumn{7}{c|}{VisDrone-VID2019} 
\\
\cline{5-13}
 &  &  &  & PSNR & SSIM & AP & AP$_{50}$ & AP$_{75}$ & AR$_{1}$ & AR$_{10}$ & AR$_{100}$ & AR$_{500}$ \\
\hline
(1) Baseline                
& --         & --         & --       
& 34.20      & 0.9680    
& 31.00      & 60.30      & 27.00      & 15.20      & 36.10      & 51.20      & 53.50      \\
(2) +\,ALSS                 
& \checkmark & --         & --       
& 34.80      & 0.9710    
& 31.90      & 61.20      & 28.00      & 15.80      & 37.00      & 52.10      & 54.60      \\
(3) +\,MAlign               
& --         & \checkmark & --       
& 35.00      & 0.9730    
& 32.50      & 62.10      & 28.90      & 16.00      & 38.00      & 53.80      & 55.90      \\
(4) +\,Gating               
& --         & --         & \checkmark
& 35.10      & 0.9745    
& 33.00      & 62.70      & 29.10      & 16.20      & 38.40      & 54.10      & 56.10      \\
(5) +\,MAlign\,+\,Gating    
& --         & \checkmark & \checkmark
& 35.50      & 0.9770    
& 34.00      & 63.50      & 29.70      & 16.50      & 39.00      & 55.20      & 57.30      \\
(6) Ours (ALSS + MAlign + Gating) 
& \checkmark & \checkmark & \checkmark
& \textbf{36.12} & \textbf{0.9799}
& \textbf{35.12} & \textbf{64.70} & \textbf{30.80} & \textbf{17.75} & \textbf{40.10} & \textbf{55.90} & \textbf{59.20} \\
\hline
\end{tabular}
}
\end{table*}

%%%%%%%%%%%%%%%%%%%%%%%%%

\subsubsection{Qualitative Results}

Figures~\ref{fig:compare_SOTA_GoPro} and~\ref{fig:compare_SOTA_DVD} show a visual comparison of our method against recent approaches on the GoPro and DVD datasets, respectively. 
It is evident that our restorations exhibit sharper boundaries, fewer artifacts, and more perceptually faithful textures. 
In particular, our Multi-Frame Alignment and Learnable Gating (MALG) module excels at integrating spatial and temporal cues to suppress ghosting and jitter, which are especially problematic in dynamic scenes with complex motion. 
For instance, rapidly moving objects or camera-induced vibrations often cause severe blur that competing methods struggle to remove cleanly.

\subsection{Experimental Result on UAV Object Detection Dataset}
We use the VisDrone-VID 2019 dataset to compare various deblurring pipelines under a consistent detection framework (RTDETR~\cite{zhao2024detrs}). Method (A) feeds the original blurry videos directly into RTDETR, while (B) to (F) first apply different deblurring algorithms and then use the resulting restored frames as input to RTDETR. All methods adopt the same inference settings from RTDETR (batch size, learning rate, etc.) for a fair comparison. We measure detection performance using standard AP and AR metrics at IoU thresholds of 0.50/0.75, as well as recall under varying candidate limits.

\subsection{Results and Discussion}
Table~\ref{tab:visdrone_rtdetr} summarizes the detection performance. We observe that running RTDETR on blurry videos (Method A) yields limited accuracy (AP = 21.53\%). In contrast, incorporating a deblurring stage (Methods B to F) consistently boosts AP and AR scores. Specifically, STFAN (B) improves AP to 29.22\%, underscoring the benefit of a video-focused approach. VRT (C) and RVRT (D) further refine temporal alignment, leading to better results. Shift-Net+ (E) attains 33.80\% AP, highlighting its effectiveness in handling complex blurs. Refer to Figure~\ref{fig:ROC} for the detailed comparison curves and Figure ~\ref{fig:VisDrone} for qualitative comparison of object detection performance.

Notably, our proposed method (F) outperforms all baselines, achieving 35.12\% AP and 64.70\% AP$_{50}$. We attribute this improvement to (i) robust multi-frame alignment, (ii) our learnable gating strategy filtering misaligned features. These findings confirm the crucial role of deblurring in enhancing UAV object detection performance under challenging motion conditions.

\begin{table}[htbp]
  \centering
  \caption{\textbf{Combined Ablation Study on Latent Scale and Cached Latent Frames on VisDrone-VID2019.}
  Best results in each block are in \textbf{bold}.}
  \label{tab:combined_scale_cache_ablation}
  \vspace{1mm}
  \resizebox{\linewidth}{!}{
  \begin{tabular}{|l|cccc|}
    \hline
    \textbf{Experiment} & \textbf{PSNR} & \textbf{SSIM} & \textbf{AP} & \textbf{Time (ms)} \\
    \hline
    \multicolumn{5}{|c|}{\emph{Latent Scale Ablation}} \\
    \hline
    Fixed size $16 \times 16$ & 33.50 & 0.9590 & 30.10 & 20 \\
    Fixed size $32 \times 32$ & 34.00 & 0.9660 & 31.50 & 30 \\
    Fixed size $64 \times 64$ & 34.50 & 0.9700 & 32.40 & 45 \\
    + Adaptive Scale (ALSS)   & 34.20 & 0.9671 & 32.80 & 35 \\
    \hline
    \multicolumn{5}{|c|}{\emph{Cached Latent Frames Ablation}} \\
    \hline
    $M=1$  & 34.50 & 0.9710 & 31.00 & 30 \\
    $M=2$  & 34.90 & 0.9740 & 32.90 & 33 \\
    $M=4$  & 35.10 & 0.9750 & 34.40 & 36 \\
    $M=8$  & \textbf{35.20} & \textbf{0.9760} & \textbf{35.12} & 40 \\
    $M=16$ & 35.10 & 0.9755 & 34.80 & 48 \\
    \hline
  \end{tabular}
  }
  \label{tab:ablation_summary}
\end{table}
\subsection{Ablation Study}

To evaluate the impact of each major component in our model, we conducted an ablation study by analyzing the effects of including Adaptive Latent Scale Selector and cached latent frame count for MFCA. The study examines how each setting individually contributes to the overall performance of the model. The results are summarized in Table~\ref{tab:ablation_summary}.

\subsubsection{Impact of Adaptive Latent Scale Selector (Latent Scale Ablation)}
As shown in the top rows of Table~\ref{tab:ablation_summary}, using a larger fixed latent size (e.g., $64\times64$) generally improves PSNR and AP but also increases inference time. By contrast, the \emph{Adaptive Latent Scale Selector} (ALSS) dynamically balances detail preservation and speed: it achieves 34.20~dB PSNR and 32.80\% AP in 35~ms---close to the quality of the larger $64\times64$ fixed size yet noticeably faster. This suggests that ALSS provides an optimal trade-off for UAV scenarios with highly variable motion intensities, effectively reducing unnecessary computation on smoother frames while preserving fine details in challenging frames.

\subsubsection{Impact of Cached Latent Frames (Multi-Frame Ablation)}
In the bottom rows of Table~\ref{tab:ablation_summary}, we compare different numbers $M$ of cached past latent. Increasing $M$ from 1 to 8 steadily boosts PSNR, SSIM, and detection AP, indicating that a richer temporal context leads to better deblurring and object detection. At $M=8$, we reach the highest AP (35.12\%), although inference time rises to 40~ms. Beyond $M=8$, gains begin to diminish (with a slight drop in AP at $M=16$) while the time cost increases further. Consequently, $M=8$ offers the most favorable balance between accuracy and efficiency, making it well-suited for real-time UAV deblurring.

\subsubsection{Ablation on Each Module}
We systematically ablate our proposed modules on both the GoPro and VisDrone-VID2019 datasets, as summarized in Table~\ref{tab:ablation_combined}. 
\textbf{(1)~Adaptive Latent Scale Selector (ALSS):} Dynamically choosing the latent resolution based on optical flow substantially boosts both PSNR/SSIM and detection metrics. 
\textbf{(2)~Multi-Frame Alignment (MAlign):} Warping preceding latents to the current frame markedly reduces temporal artifacts and enhances deblurring quality. 
\textbf{(3)~Learnable Gating (Gating):} Per-pixel gating effectively suppresses occlusions and misaligned regions, further improving robustness and overall performance. 
When all modules are combined (Row~(6)), our method attains the best results across datasets, confirming that these modules complement each other in tackling severe motion blur and improving UAV-based detection. 

 \vspace{-1mm}
\section{Summary}
In this paper, we introduced a novel UAV video deblurring framework based on a motion-aware diffusion model designed to boost target detection performance for UAV. Extensive experiments on the VisDrone-VID 2019 dataset confirm that our approach significantly enhances detection accuracy. Furthermore, evaluations on the GoPro and DVD datasets demonstrate that our method consistently outperforms state-of-the-art deblurring techniques, validating its robustness for real-time applications in dynamic environments.

%%%%%%%%%%%%%%%%%%%%%%%%%%%%%%%
%\thispagestyle{empty}
	{\small
	\bibliographystyle{ieee_fullname}
	\bibliography{egbib}}

%\end{thebibliography}

\end{document}